\documentclass{article}
\usepackage{spconf,amsmath,graphicx,hyperref}
\usepackage{textcomp}
\usepackage{xcolor}
\usepackage{subcaption}
\usepackage{multirow}
\usepackage{threeparttable}
\usepackage{algorithmic}
\usepackage{graphicx}
\usepackage{amssymb}

\usepackage{xspace} %
\def\onedot{\ifx\@let@token.\else.\null\fi\xspace} 
\def\eg{\emph{e.g}\onedot}

\def\etc{\emph{etc}\onedot}

\makeatother 

\usepackage{booktabs}  
\usepackage{amsmath}

\usepackage{enumitem}
\usepackage{colortbl}
\definecolor{bgreen}{RGB}{0,170,0}
\definecolor{bred}{RGB}{220,0,0}
\definecolor{mydarkblue}{RGB}{0,0,150}
\definecolor{Grey}{gray}{0.85}
\definecolor{darkGrey}{gray}{0.66}
\definecolor{Gray}{rgb}{0.88,0.88,0.96}
\definecolor{Purple}{rgb}{0.6,0.6,0.96}
\hypersetup{
    colorlinks=true,
    linkcolor=bred,
    citecolor=mydarkblue,
    filecolor=bred,
    urlcolor=mydarkblue
}

\usepackage{wrapfig}

\usepackage{xcolor}
\definecolor{y}{RGB}{255, 250, 205}
\definecolor{p}{RGB}{245, 234, 240}
\definecolor{b}{RGB}{224, 255, 255}
\usepackage{algorithm}
\usepackage{makecell}
\usepackage{float}  

\definecolor{PineGreen}{HTML}{008B00}
\definecolor{BrickRed}{HTML}{B22222}

\usepackage[textsize=tiny]{todonotes}
\usepackage{bbm} 
\usepackage{subcaption} 
\usepackage{graphicx} 
\usepackage{caption}  
\usepackage{multirow} 
\usepackage{pifont}  
\usepackage{svg} 
\usepackage{listings}   

\title{QA-ReID: Quality-Aware Query-Adaptive Convolution Leveraging Fused Global and Structural Cues for Clothes-Changing ReID}
\name{Yuxiang Wang\textsuperscript{1}, Kunming Jiang\textsuperscript{2}, Tianxiang Zhang\textsuperscript{3}, Ke Tian\textsuperscript{4}, Gaozhe Jiang\textsuperscript{5}*\thanks{Yuxiang Wang (yuxiang.wang0476@gmail.com), Kunming Jiang (fnwjkm@gmail.com), Tianxiang Zhang (tz534@nyu.edu), Ke Tian (ketian2@illinois.edu). Corresponding author: Gaozhe Jiang (e0945601@u.nus.edu). }}
\address{\textsuperscript{1}School of Business, Stevens Institute of Technology, New York City, the United States, 10018\\
\textsuperscript{2}Electrical and Computer Engineering, University of California, San Diego, La Jolla, CA, USA, 92093\\
\textsuperscript{3}Tandon School of Engineering, New York University, New York, USA, 10003\\
\textsuperscript{4}Department of Electrical and Computer Engineering,\\ University of Illinois at Urbana-Champaign, Champaign, USA, 61820\\
\textsuperscript{5}Institute of Operations Research and Analytics, National University of Singapore, Singapore, 119007
}

\begin{document}
\ninept
\maketitle
\begin{abstract}
Unlike conventional person re-identification (ReID), clothes-changing ReID (CC-ReID) presents severe challenges due to substantial appearance variations introduced by clothing changes. In this work, we propose the Quality-Aware Dual-Branch Matching (QA-ReID), which jointly leverages RGB-based features and parsing-based representations to model both global appearance and clothing-invariant structural cues. These heterogeneous features are adaptively fused through a multi-modal attention module. At the matching stage, we further design the Quality-Aware Query Adaptive Convolution (QAConv-QA), which incorporates pixel-level importance weighting and bidirectional consistency constraints to enhance robustness against clothing variations. Extensive experiments demonstrate that QA-ReID achieves state-of-the-art performance on multiple benchmarks, including PRCC, LTCC, and VC-Clothes, and significantly outperforms existing approaches under cross-clothing scenarios.
\end{abstract}


\section{Introduction}
Person re-identification (ReID)\cite{short-1,short-6} is a core computer vision task that focuses on retrieving and matching pedestrian identities across non-overlapping camera views.
It has critical applications in intelligent surveillance, public safety, and missing-person search, making it a focal point of research in both academia and industry. The core challenge of ReID lies in learning discriminative yet robust feature representations that can withstand environmental variations such as illumination, pose, viewpoint, and occlusion. In clothing-changing scenarios, however, conventional methods rely heavily on appearance cues, including color, texture, and style. When a pedestrian changes clothes, these features undergo substantial alterations, often leading to drastic performance drops or even complete failure. Therefore, the primary challenge in clothes-changing person ReID (CC-ReID)\cite{my-iotj,long-1,local-2} is to mitigate the interference caused by clothing variations and instead exploit more stable biometric cues, such as body shape, gait, hairstyle, or carried belongings.
\begin{figure}[tbp]
\centerline{\includegraphics[width=0.42\textwidth]{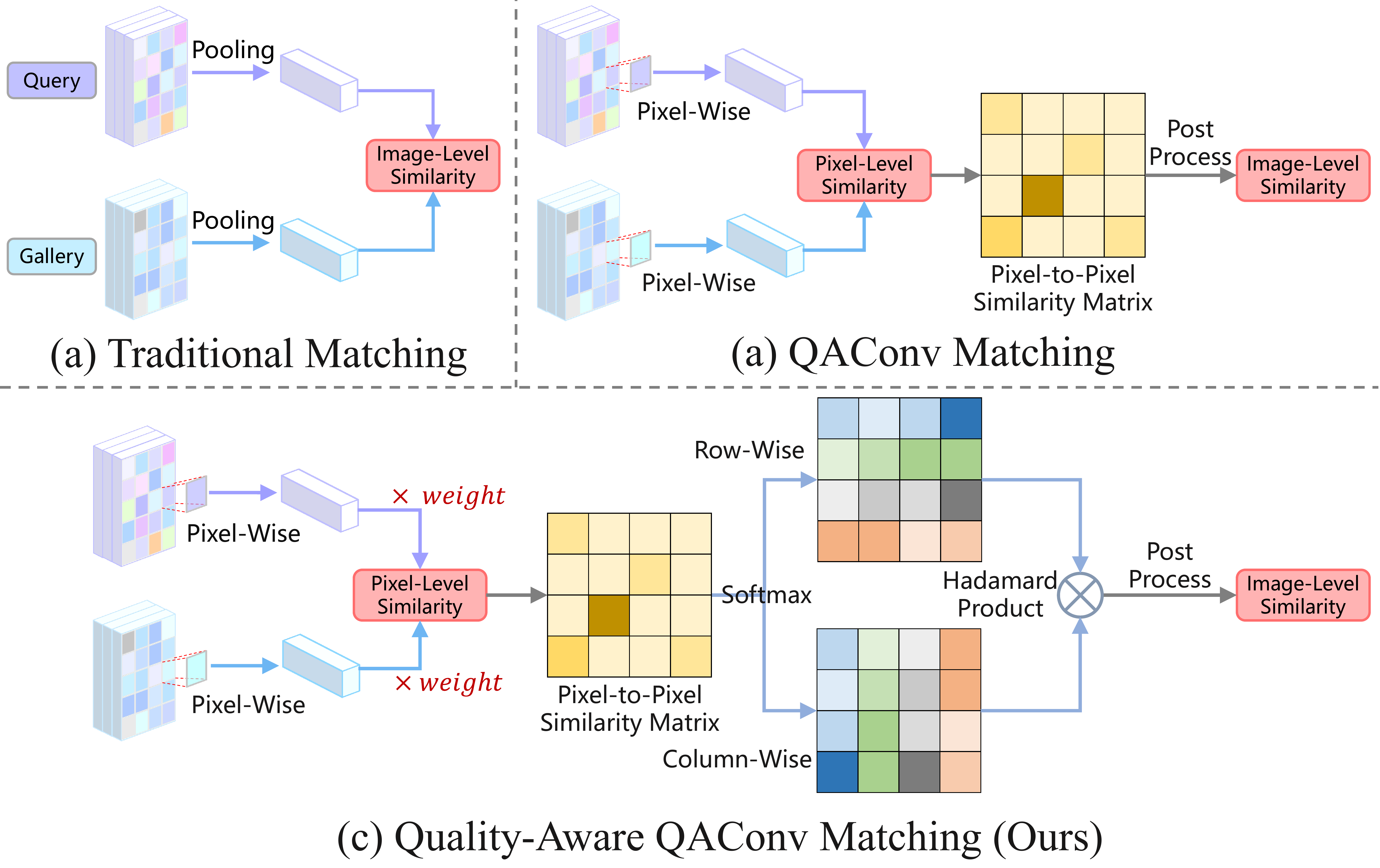}}
 
\caption{ (a) The conventional global matching. (b) Local matching using QAConv. (c) Our proposed QAConv-QA.}
\label{teaser}
 
\end{figure}

To address this challenge, various research directions have been explored. One line of approaches leverages inherent human structural information, extracting body contours\cite{prcc,3d,3dwacv} and skeletal keypoints\cite{ltcc} via human parsing\cite{my-air} or pose estimation\cite{fsam} to mitigate the influence of clothing. Another line of work focuses on designing more powerful deep networks, such as incorporating Generative Adversarial Networks (GANs)\cite{gan-cloth-CASE-Net,gan-cloth2-IS-GAN} or feature disentanglement techniques to separate identity-related features from clothing attributes. Nevertheless, despite their progress, these methods remain insufficient. Region-based methods are vulnerable to pose estimation errors and occlusions, while generative approaches often suffer from low-quality or distorted synthetic details. More importantly, most existing frameworks provide limited exploration in feature fusion and remain under-optimized at the matching stage (see \textbf{Fig.~\ref{teaser}-a}).


To overcome these limitations, we propose a novel Quality-Aware Dual-Branch Matching (QA-ReID) framework, which begins with a parallel dual-branch network: one branch directly processes raw RGB images to capture global appearance, while the other extracts clothing-invariant biometric structural features via a human parsing module. To achieve deep integration of the two modalities, we design a multi-modal attention module that enhances the discriminability of the fused features.

\begin{figure*}[htbp]
\centerline{\includegraphics[width=0.82\textwidth]{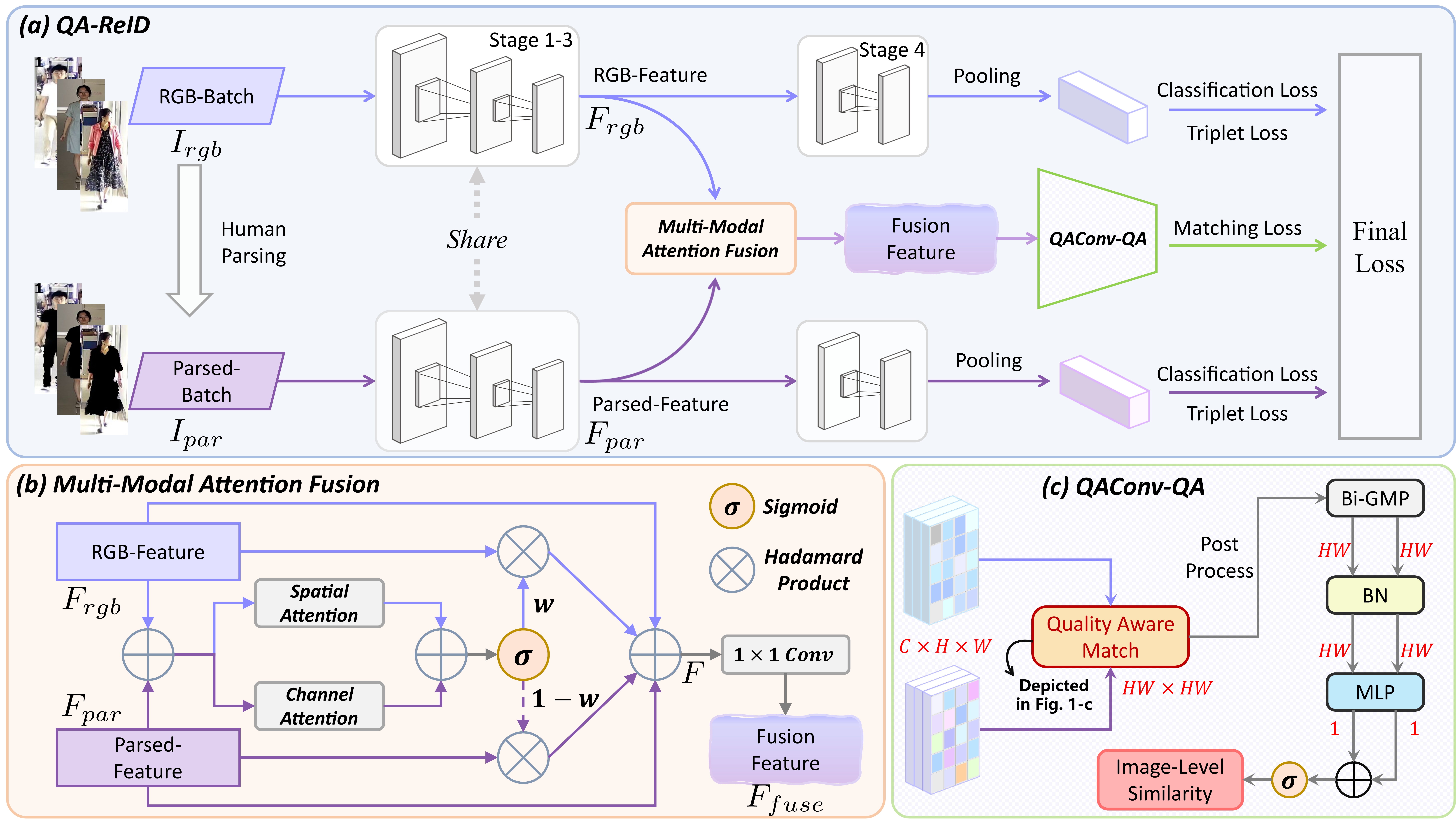}}
 
\caption{Illustration of our QA-ReID, which takes both RGB images and clothing-invariant representations obtained from human parsing as inputs. Then, a multi-modal attention fusion module is employed to combine the complementary features, and a quality-aware query adaptive convolution module (QAConv-QA) is introduced to compute similarity scores between images. }
\label{main}
 
\end{figure*}

Based on the fused features, we adopt the Query-Adaptive Convolution (QAConv) \cite{QAConv} (see \textbf{Fig.~\ref{teaser}-b}) in the matching stage of QA-ReID. Conventional matching strategies typically treat all feature regions equally, overlooking that some regions contain more discriminative identity cues while others are easily affected by \textbf{clothing variations} or background noise. QAConv alleviates this issue by emphasizing pixel-level similarities, enabling the model to focus on informative regions. Building upon this, we introduce two key improvements and propose the Quality-Aware Query-Adaptive Convolution (QAConv-QA, see \textbf{Fig.~\ref{teaser}-c}): (1) emphasizing key human regions through enhanced pixel-level weighting and (2) explicitly modeling bidirectional matching to improve stability and accuracy.


The main contributions of this work are summarized as follows: \ding{182} We propose a dual-branch framework tailored for clothes-changing person Re-ID, which effectively fuses RGB-based and clothing-invariant features via an attention mechanism. \ding{183} We design the Quality-Aware Query Adaptive Convolution (QAConv-QA), which introduces pixel-level importance weighting and explicit bidirectional matching, significantly improving upon the original QAConv. \ding{184} Extensive experiments on multiple CC-ReID benchmarks demonstrate that our method achieves superior performance and sets new state-of-the-art results under clothing-changing conditions.

\section{Method}

To address the appearance shift caused by clothing variations in clothes-changing person Re-ID, we propose an end-to-end framework, QA-ReID, which integrates dual-branch feature extraction with quality-aware pixel-level matching, as illustrated in \textbf{Fig.~\ref{main}}. In the following, we provide a detailed description of each component of QA-ReID.

\begin{table*}[htbp]
\renewcommand{\arraystretch}{1.0}
\renewcommand\tabcolsep{5pt}
\centering
\small
\caption{\centering Comparison With State-of-the-art Methods On PRCC, LTCC and VC-Clothes.}\label{tab1}
 
\begin{threeparttable}
\begin{tabular}{c|c|cccc|cccc|cccc}
\toprule[1.1pt]
\multirow{3}{*}{Methods} & \multirow{3}{*}{Venue} & \multicolumn{4}{c|}{PRCC}                                                           & \multicolumn{4}{c|}{LTCC}                                                    & \multicolumn{4}{c}{VC-Clothes}                                                     \\ 
  \cmidrule(lr){3-6} \cmidrule(lr){7-10} \cmidrule(lr){11-14} 
                         &                        & \multicolumn{2}{c|}{SC}                             & \multicolumn{2}{c|}{CC}       & \multicolumn{2}{c|}{General}                       & \multicolumn{2}{c|}{CC}       & \multicolumn{2}{c|}{General}                       & \multicolumn{2}{c}{CC}        \\ 
  \cmidrule(lr){3-4} \cmidrule(lr){5-6} \cmidrule(lr){7-8} \cmidrule(lr){9-10}  \cmidrule(lr){11-12} \cmidrule(lr){13-14}
                         &                        & Top-1          & \multicolumn{1}{c|}{mAP}           & Top-1         & mAP           & Top-1         & \multicolumn{1}{c|}{mAP}           & Top-1         & mAP           & Top-1         & \multicolumn{1}{c|}{mAP}           & Top-1         & mAP           \\ 
\cmidrule(lr){1-1} \cmidrule(lr){2-2}    \cmidrule(lr){3-4} \cmidrule(lr){5-6} \cmidrule(lr){7-8} \cmidrule(lr){9-10}  \cmidrule(lr){11-12} \cmidrule(lr){13-14}
PCB\cite{pcb}                      & ECCV18                 & 99.8           & \multicolumn{1}{c|}{97.0}          & 41.8          & 38.7          & 65.1          & \multicolumn{1}{c|}{30.6}          & 23.5          & 10.0          & 87.6          & \multicolumn{1}{c|}{74.6}          & 62.0          & 62.3          \\
TransReID\cite{transreid}                & ICCV21                 & 97.3           & \multicolumn{1}{c|}{95.9}          & 47.1          & 49.3          & 73.8          & \multicolumn{1}{c|}{39.4}          & 35.7          & 18.8          & 90.5          & \multicolumn{1}{c|}{80.1}          & 70.0          & 71.8          \\ 

\cmidrule(lr){1-1} \cmidrule(lr){2-2}    \cmidrule(lr){3-4} \cmidrule(lr){5-6} \cmidrule(lr){7-8} \cmidrule(lr){9-10}  \cmidrule(lr){11-12} \cmidrule(lr){13-14}
GI-ReID\cite{gait}                  & CVPR22                 & 80.0           & \multicolumn{1}{c|}{-}             & 30.3          & -             & 63.2          & \multicolumn{1}{c|}{29.4}          & 23.7          & 10.4          & -             & \multicolumn{1}{c|}{-}             & 57.8          & 64.5          \\
CAL\cite{cal}                      & CVPR22                 & \textbf{100.0} & \multicolumn{1}{c|}{99.8}          & 55.2          & 55.8          & 74.2          & \multicolumn{1}{c|}{40.8}          & 40.1          & 18.0          & 92.9          & \multicolumn{1}{c|}{87.2}          & 55.2          & 55.8          \\
DCR-ReID\cite{dcr-reid}                 & TCSVT23                & \textbf{100.0} & \multicolumn{1}{c|}{99.7}          & 57.2          & 57.4          & 76.1          & \multicolumn{1}{c|}{\textbf{42.3}} & 41.1          & 20.4          & \textbf{-}    & \multicolumn{1}{c|}{-}             & -             & -             \\
CLIP3DReID\cite{clip3dreid}               & CVPR24                 & -              & \multicolumn{1}{c|}{-}             & 60.6          & 59.3          & -             & \multicolumn{1}{c|}{-}             & 42.1          & \textbf{21.7} & -             & \multicolumn{1}{c|}{-}             & -             & -             \\
MCSC\cite{mcsc}                     & TIP24                  & 99.8           & \multicolumn{1}{c|}{99.8}          & 57.8          & 57.3          & 73.9          & \multicolumn{1}{c|}{40.2}          & 42.2          & 19.4          & 93.2          & \multicolumn{1}{c|}{87.9}          & 83.3          & 83.2          \\
RFFR-Net\cite{rffr-net}                 & Inffus25               & \textbf{100.0} & \multicolumn{1}{c|}{\textbf{99.9}} & 56.6          & 57.5          & 76.0          & \multicolumn{1}{c|}{41.4}          & 41.4          & 19.8          & \textbf{-}    & \multicolumn{1}{c|}{\textbf{-}}    & -             & -             \\
\cmidrule(lr){1-1} \cmidrule(lr){2-2}    \cmidrule(lr){3-4} \cmidrule(lr){5-6} \cmidrule(lr){7-8} \cmidrule(lr){9-10}  \cmidrule(lr){11-12} \cmidrule(lr){13-14}
QA-ReID                  & -                      & \textbf{100.0} & \multicolumn{1}{c|}{99.5}          & \textbf{64.1} & \textbf{61.2} & \textbf{76.2} & \multicolumn{1}{c|}{41.3}          & \textbf{42.9} & 21.3          & \textbf{95.1} & \multicolumn{1}{c|}{\textbf{90.2}} & \textbf{86.3} & \textbf{86.1} \\ 
\bottomrule[1.1pt]
\end{tabular}

\end{threeparttable}
 
\end{table*}

\subsection{Feature Extraction and Fusion}

At the input stage, we employ a human parsing network $P(\cdot)$ to parse each image $I \in \mathbb{R}^{3 \times H' \times W'}$, producing a segmentation map $M \! = \! P(I) \in {\{1,\dots,K\}}^{H' \times W'}$, where $K$ denote the number of semantic classes (\eg, hair, face, \etc). A binary human mask $M_{body} \! \in \! {\{0,1\}}^{H' \times W'} $ is then constructed from $M$ to retain identity-related regions such as the head and limbs. Based on this mask, we obtain the clothing-invariant image $I_{par} = I \odot M_{body}$. The original image $I_{rgb}=I$ and the parsed image $I_{par}$ are fed into parallel branches to extract feature representations $F_{rgb}  \in \mathbb{R}^{C\times H\times W} $ and $F_{par}  \in \mathbb{R}^{C\times H\times W}$, respectively. It is worth noting that these features are extracted after the third stage of the backbone, which in this work is ResNet50.

To integrate the two complementary types of features, we design a multi-modal attention fusion module. After concatenating the two feature maps, we apply a channel attention branch and a spatial attention branch to generate the corresponding attention maps, which are then combined multiplicatively to form a joint attention map $\omega \in [0,1]^{C\times H\times W}$. The weight map $\omega$ is adaptively learned by the network and reflects the relative importance of RGB features and parsing features across both channel and spatial dimensions. Guided by the attention map, the two features are fused as follows:  
\begin{equation}
F_{mix} = \omega \odot F_{rgb} + (1-\omega) \odot F_{par},
\label{eq:f_mix}
\end{equation}
Finally, we integrate three components---$F_{rgb}$, $F_{par}$, and $F_{mix}$---to obtain the combined feature representation:  
\begin{equation}
F = F_{rgb} + F_{par} + F_{mix},
\label{eq:f_combine}
\end{equation}
followed by a $1\times1$ convolution for feature compression and refinement, resulting in the final fused representation $F_{fuse}$. This design preserves the independent representations of the RGB and parsing branches while allowing the attention weights $\omega$ to adaptively balance their contributions at both the pixel and channel levels, thereby producing more robust and discriminative fused features.


\subsection{Quality-Aware Pixel-Level Matching}

Based on the fused features, we introduce QAConv-QA, a quality-aware pixel-level matching mechanism for measuring similarity between image pairs. Compared with the original QAConv, our QAConv-QA enhances pixel similarity computation by: (1) emphasizing key human regions via refined pixel-level weighting, and (2) explicitly modeling bidirectional matching to improve stability and accuracy.

\noindent\textbf{Pixel Weight.} For the fused representation $F_{fuse} \in \mathbb{R}^{C\times H\times W}$, each pixel is assigned a quality weight that reflects its identity relevance, the weight matrix is denoted as $Q \in \mathbb{R}^{ H\times W}$: 
\begin{equation}
\begin{aligned}
    \bar{Q}_{i,j} & = \frac{1}{k^2} \sum_{i' = i \cdot k}^{i \cdot k + k - 1} \sum_{j' = j \cdot k}^{j \cdot k + k - 1} {M_{\mathrm{body}}}_{ i', j'}, \\
    Q_{i,j} & = \frac{\exp(\bar{Q}_{i,j})}{\sum_{h=1}^{H}\sum_{w=1}^{ W}\exp(\bar{Q}_{h,w})},
\end{aligned}
\end{equation}
where $k = \lfloor  \frac{H'}{H} \rfloor  = \lfloor  \frac{W'}{W} \rfloor $. According to Eq.~(3), pixels located on critical body parts are assigned greater weights. Based on $Q$, the weighted similarity between a pixel pair \((f^1_{i_1,j_1}, f^2_{i_2,j_2})\) from $F_{fuse}^1$ and $F_{fuse}^2$ is defined as:
\begin{equation}
\text{sim}^1(f^1_{i_1, j_1}, f^2_{i_2,j_2}) = Q^1_{i_1,j_1} \cdot Q^2_{i_2,j_2} \cdot \rho(f^1_{i_1, j_1}, f^2_{i_2,j_2}).
\end{equation}
where $\rho$ denote the cosine similarity.

\noindent\textbf{Bidirectional Matching.} We further introduce a bidirectional matching constraint:
\begin{equation}
\begin{aligned}
& \text{sim}^2(f^1_{i_1, j_1}, f^2_{i_2,j_2})   =   \bar \rho(f^1_{i_1, j_1}|f^2_{i_2,j_2})  \cdot  \bar \rho(f^2_{i_2,j_2}|f^1_{i_1, j_1}), \\
& \bar \rho(f^1_{i_1, j_1}|f^2_{i_2,j_2}) = \frac{\exp({\text{sim}^1(f^1_{i_1, j_1}, f^2_{i_2,j_2})})}{\sum_{h=1}^{H}\sum_{w=1}^{W}\exp({\text{sim}^1(f^1_{h, w}, f^2_{i_2,j_2})})}.
\end{aligned}
\end{equation}
This design ensures that only pixel pairs with high quality and bidirectional consistency make significant contributions to the final similarity score, thereby effectively reducing mismatches. During the post process, the pixel-level matching scores are then aggregated into an image-level similarity score via bidirectional global maximum pooling (Bi-GMP), followed by batch normalization,  an MLP layer and a sigmoid layer.

\subsection{Joint Loss Function}

We design a multi-task joint loss function that combines identity classification, triplet, and matching losses. Identity classification loss is applied to the global features from both RGB and parsed branches using cross-entropy. Triplet loss is introduced to enforce intra-class compactness and inter-class separability in the embedding space. To optimize QAConv-QA, we employ a pairwise matching loss based on binary cross-entropy, which supervises the predicted similarity scores between image pairs. For a mini-batch of size \( B \), the loss is defined as:
\begin{equation}
L_{\text{match}} = -\frac{1}{B^2} \sum_{i=1}^B \sum_{j=1}^B \big[ y_{ij} \log p_{ij} + (1-y_{ij}) \log (1-p_{ij}) \big],
\end{equation}
where \( y_{ij} \) indicates whether two images belong to the same identity, \( p_{ij} \) denote the output of the post process. The overall objective is given by:
\begin{equation}
L = (L_{\text{cls}}^{\text{rgb}} + L_{\text{cls}}^{\text{par}}) + (L_{\text{tri}}^{\text{rgb}} + L_{\text{tri}}^{\text{par}}) + L_{\text{match}}.
\end{equation}
By jointly optimizing all objectives, the framework achieves effective synergy across feature extraction, fusion, and matching, ultimately benefit clothing-changing conditions.

\section{Experiments}

\subsection{Experimental Settings}

We evaluate the effectiveness of the proposed method on three representative datasets. The PRCC dataset contains 221 identities and 33,698 images, including both same-clothing and clothing-changing samples. The LTCC dataset includes 152 identities with 17,119 images, emphasizing long-term temporal variations with multiple clothing changes. The VC-Clothes dataset is a synthetic benchmark generated by a virtual engine, containing 512 identities and 19,060 images to simulate multi-scene and multi-clothing scenarios. For evaluation, we adopt mean Average Precision (mAP) and Cumulative Matching Characteristics (CMC) at Top-1 under three testing protocols: general setting, clothing-changing (CC) setting, and same-clothing (SC) setting.

For implementation, we adopt the PyTorch framework and conduct training on a single NVIDIA RTX 3090 GPU. Input images are resized to \(384 \times 192\), with data augmentation including random flipping, cropping, and erasing\cite{random-eraser}. ResNet-50 pretrained on ImageNet is employed as the backbone for both RGB and parsing branches, following QAConv \cite{QAConv}, we also conduct QAConv-QA after the third stage of the ResNet-50. We use Adam as the optimizer with an initial learning rate of \(3.5 \times 10^{-4}\), decayed by 0.1 every 40 epochs, for a total of 100 epochs. The batch size is set to 32.

\subsection{Comparison with State-of-the-Art Methods}

To comprehensively validate the effectiveness of our method, we compare QA-ReID with two conventional person Re-ID methods and six representative approaches designed for clothing-changing scenarios. As shown in Table I, traditional methods perform reasonably well under the same-clothing (SC) setting but experience a significant drop in both accuracy and mAP under the clothing-changing (CC) setting, indicating the limitation of relying solely on clothing appearance features. In contrast, CC-ReID approaches that leverage parsing, feature disentanglement, or multimodal strategies achieve better performance by alleviating the impact of clothing variations, outperforming conventional methods overall.

Moreover, our proposed QA-ReID consistently achieves the best results under the clothing-changing setting on both PRCC and VC-Clothes. Specifically, it attains 64.1\% Top-1 and 61.2\% mAP on PRCC, and 86.3\% Top-1 and 86.1\% mAP on VC-Clothes, clearly surpassing existing methods. On LTCC, QA-ReID also achieves the best Top-1 score of 42.9\% under the clothing-changing protocol. These results highlight the advantage of our quality-aware matching mechanism, which substantially improves robustness and recognition performance in cross-clothing scenarios.

\begin{table}[tbp]
\renewcommand{\arraystretch}{1.0}
\renewcommand\tabcolsep{1.5pt}
\centering
\small
\caption{\centering The ablation study of QA-ReID. QA refers to the utilization of QAConv-QA.}\label{tab2}
\vspace{-0.5em}
\begin{threeparttable}
\begin{tabular}{b{0.5cm}<{\centering}|b{0.8cm}<{\centering}b{0.8cm}<{\centering}b{0.8cm}<{\centering}b{0.8cm}<{\centering}|b{1.0cm}<{\centering}b{1.0cm}<{\centering}|b{1.0cm}<{\centering}b{1.0cm}<{\centering}}
\toprule[1.1pt]
\multirow{2}{*}{ID} & \multirow{2}{*}{RGB} & \multirow{2}{*}{Parse} & \multirow{2}{*}{Fusion} & \multirow{2}{*}{QA} & \multicolumn{2}{c|}{PRCC}     & \multicolumn{2}{c}{LTCC}      \\  \cmidrule(lr){6-7} \cmidrule(lr){8-9}
                    &                      &                        &                         &                            & Top-1        & mAP           & Top-1        & mAP           \\ 
  \cmidrule(lr){1-1} \cmidrule(lr){2-5}                 \cmidrule(lr){6-7} \cmidrule(lr){8-9} 
1                   & \checkmark          &                        &                         &                            & 55.0          & 52.0          & 34.0          & 16.5          \\
2                   & \checkmark          & \checkmark            &                         &                            & 58.2          & 55.3          & 37.5          & 18.2          \\
3                   & \checkmark          & \checkmark            &                         & \checkmark                & 62.1          & 59.8          & 41.0          & 19.7          \\
4                   & \checkmark          & \checkmark            & \checkmark             &                            & 61.0          & 58.4          & 40.7          & 20.0          \\
5                   & \checkmark          & \checkmark            & \checkmark             & \checkmark                & \textbf{64.1} & \textbf{61.2} & \textbf{42.9} & \textbf{21.3} \\ 
\bottomrule[1.1pt]
\end{tabular}
    
\end{threeparttable}
 
\end{table}

\begin{table}[tbp]
\renewcommand{\arraystretch}{1.0}
\renewcommand\tabcolsep{0.1pt}
\centering
\small
\caption{\centering The ablation study of our QAConv-QA. }\label{tab3}
 \vspace{-0.5em}
\begin{threeparttable}
\begin{tabular}{b{0.5cm}<{\centering}|b{1.1cm}<{\centering}b{1.1cm}<{\centering}b{1.cm}<{\centering}b{0.8cm}<{\centering}|b{1.0cm}<{\centering}b{1.0cm}<{\centering}|b{1.0cm}<{\centering}b{1.0cm}<{\centering}}
\toprule[1.1pt]
\multirow{2}{*}{ID} & \multirow{2}{*}{Fusion} & \multirow{2}{*}{QAConv} & \multirow{2}{*}{QW} & \multirow{2}{*}{BM} & \multicolumn{2}{c|}{PRCC} & \multicolumn{2}{c}{LTCC} \\  
\cmidrule(lr){6-7} \cmidrule(lr){8-9}
& & & & & Top-1 & mAP & Top-1 & mAP \\  
\cmidrule(lr){1-1} \cmidrule(lr){2-5} \cmidrule(lr){6-7} \cmidrule(lr){8-9}
4 & \checkmark &  &  &  & 61.0 & 58.4 & 40.7 & 20.0 \\
6 & \checkmark & \checkmark &  &  & 61.4 & 59.1 & 41.1 & 19.9 \\
7 & \checkmark & \checkmark & \checkmark &  & 63.0 & 60.5 & 41.8 & 20.5 \\
5 & \checkmark & \checkmark & \checkmark & \checkmark & \textbf{64.1} & \textbf{61.2} & \textbf{42.9} & \textbf{21.3} \\
\bottomrule[1.1pt]
\end{tabular}
    
\end{threeparttable}
  \vspace{-0.5em}
\end{table}

\subsection{Ablation Study}
 
To validate the effectiveness of each component in the proposed framework, \textbf{Tab.~\ref{tab2}} presents the analysis of different branches and the QAConv-QA module. As shown, using only the RGB branch yields limited performance. Introducing the Parse branch leads to consistent improvements on both datasets. Adding the Fusion branch further enhances the results by better integrating complementary features. Finally, equipping the model with the QAConv-QA module achieves the best performance. These results clearly demonstrate the effectiveness of multi-branch feature learning and the proposed quality-aware matching mechanism.  

Furthermore, \textbf{Tab.~\ref{tab3}} reports ablation results on the key designs within QAConv-QA, also evaluated under the CC setting. The results indicate that directly introducing QAConv on the fusion branch brings modest improvements. Incorporating pixel weighting (PW) or bidirectional matching (BM) individually yields additional gains, while combining both achieves the most significant improvement. The complete QAConv-QA module ultimately obtains the best results on PRCC and LTCC under clothing-changing conditions, verifying the critical role of these two design choices in enhancing the robustness of pixel-level matching.

\subsection{Visualization}

\begin{figure}[tbp]
\centerline{\includegraphics[width=0.45\textwidth]{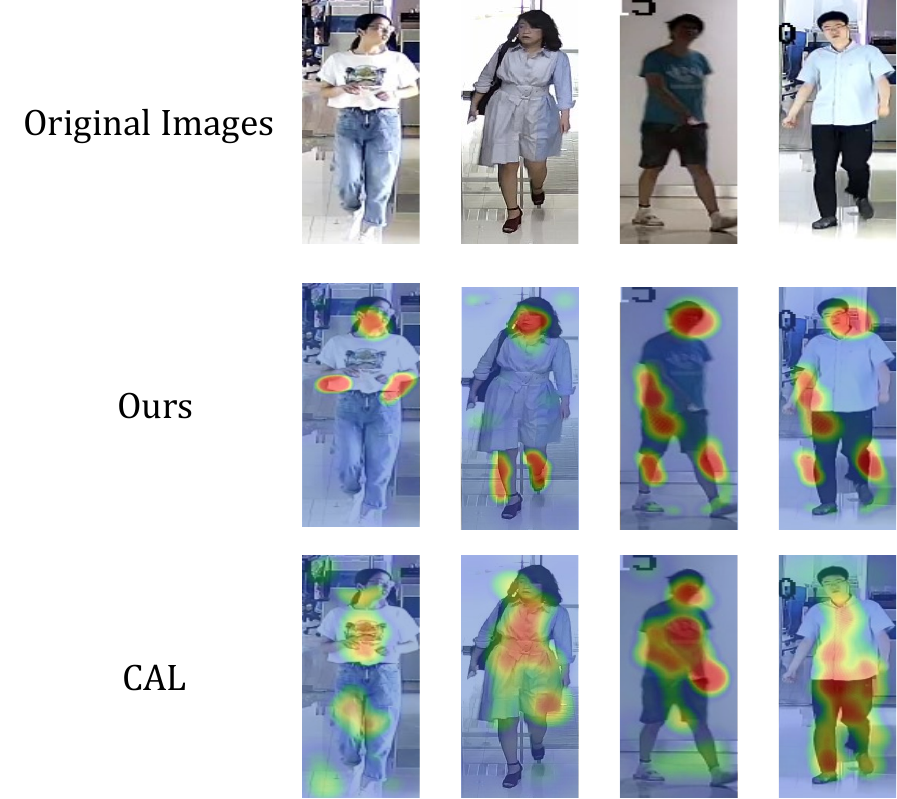}}
 
\caption{Visualization of attention maps on LTCC dataset, comparing our method with CAL.}
\label{fig3}
  \vspace{-1.0em}
\end{figure}

To further illustrate the discriminative capability of our model under clothing-changing scenarios, we visualize the attention regions using heatmaps. As shown in \textbf{Fig.~\ref{fig3}}, compared with the open-source method CAL, our approach more consistently focuses on identity-related key regions such as the head and limbs, rather than clothing-sensitive areas. This demonstrates that the proposed quality-aware matching mechanism effectively suppresses clothing interference and enables the model to achieve stronger robustness and generalization in cross-clothing scenarios.

\section{Conclusion}

In this paper, we proposed a novel framework for clothing-changing person re-identification that integrates dual-branch feature extraction with quality-aware matching. By fusing RGB and parsing features and introducing the QAConv-QA module, our model effectively suppresses clothing interference while extracting stable identity features. Extensive experiments on three challenging CC-ReID datasets demonstrate that the proposed method achieves superior performance in terms of Rank-1 accuracy and mAP, validating its robustness and generalization ability under clothing-changing conditions. In future work, we will further investigate lightweight designs and cross-modal extensions to promote the practical applications of clothing-changing person re-identification.

\bibliographystyle{IEEEbib}
\bibliography{strings,refs}

\end{document}